\documentclass[letterpaper]{article} 
\usepackage{aaai25}  
\usepackage{times}  
\usepackage{helvet}  
\usepackage{amsmath,amsfonts}
\usepackage{amssymb}
\usepackage{tabularx}
\usepackage{booktabs}
\usepackage{tcolorbox}
\usepackage{multirow}
\usepackage{algorithm}
\usepackage{bbding}
\usepackage{algorithmic}
\usepackage{courier}  
\usepackage[hyphens]{url}  
\usepackage{graphicx} 
\usepackage{natbib}  
\usepackage{caption} 
\usepackage{algorithm}
\usepackage{algorithmic}

\usepackage{newfloat}
\usepackage{listings}
\DeclareCaptionStyle{ruled}{labelfont=normalfont,labelsep=colon,strut=off} 
\floatstyle{ruled}
\newfloat{listing}{tb}{lst}{}
\floatname{listing}{Listing}
 \title{Multi-Level Optimal Transport for Universal Cross-Tokenizer Knowledge Distillation on Language Models}
\author {
    Xiao Cui\textsuperscript{\rm 1}\equalcontrib,
    Mo Zhu\textsuperscript{\rm 2}\equalcontrib,
    Yulei Qin\textsuperscript{\rm 3},
    Liang Xie\textsuperscript{\rm 2,4},
    Wengang Zhou\textsuperscript{\rm 1},
    Houqiang Li\textsuperscript{\rm 1}\thanks{Corresponding author: Houqiang Li.}
}
\affiliations {
    \textsuperscript{\rm 1}University of Science and Technology of China
    \textsuperscript{\rm 2}Zhejiang University\\
    \textsuperscript{\rm 3}Tencent YouTu Lab  \textsuperscript{\rm 4}Zhejiang University of Technology \\
    cuixiao@mail.ustc.edu.cn, 
         \{zhwg,lihq\}@ustc.edu.cn, mozhu@zju.edu.cn, yuleiqin@tencent.com, lilydedbb@gmail.com
}

\usepackage{bibentry}

\begin{document}

\maketitle

\begin{abstract}
Knowledge distillation (KD) has become a prevalent technique for compressing large language models (LLMs). Existing KD methods are constrained by the need for identical tokenizers (i.e., vocabularies) between teacher and student models, limiting their versatility in handling LLMs of different architecture families.
In this paper, we introduce the Multi-Level Optimal Transport (MultiLevelOT), a novel approach that advances the optimal transport for universal cross-tokenizer knowledge distillation.
Our method aligns the logit distributions of the teacher and the student at both token and sequence levels using diverse cost matrices, eliminating the need
for dimensional or token-by-token correspondence.
At the token level, MultiLevelOT integrates both global and local information by jointly optimizing all tokens within a sequence to enhance robustness.
At the sequence level, we efficiently capture complex distribution structures of logits via the Sinkhorn distance, which approximates the Wasserstein distance for divergence measures.
Extensive experiments on tasks such as extractive QA, generative QA, and summarization demonstrate that the MultiLevelOT outperforms state-of-the-art cross-tokenizer KD methods under various settings.
Our approach is robust to different student and teacher models across model families, architectures, and parameter sizes. 
Codes and models are available at \url{https://github.com/2018cx/Multi-Level-OT}.
\end{abstract}

\section{Introduction}

Large language models (LLMs) such as LLaMA~\cite{LLaMA,LLaMA2,meta2024}, Mistral~\cite{mistral} and Qwen~\cite{qwen1,qwen2}
have set state-of-the-art (SOTA) records on various natural language processing (NLP) tasks.
While the scaling laws of LLMs have driven the development of larger models with billions of parameters,
their substantial sizes pose significant challenges to deployment under resource-constrained environments.
To address this issue, knowledge distillation (KD) has emerged as a cost-efficient technique
for its ability to distill smaller models that maintain competitive performance.

\begin{figure}[tbp]
\begin{center}
\includegraphics[width=\linewidth]{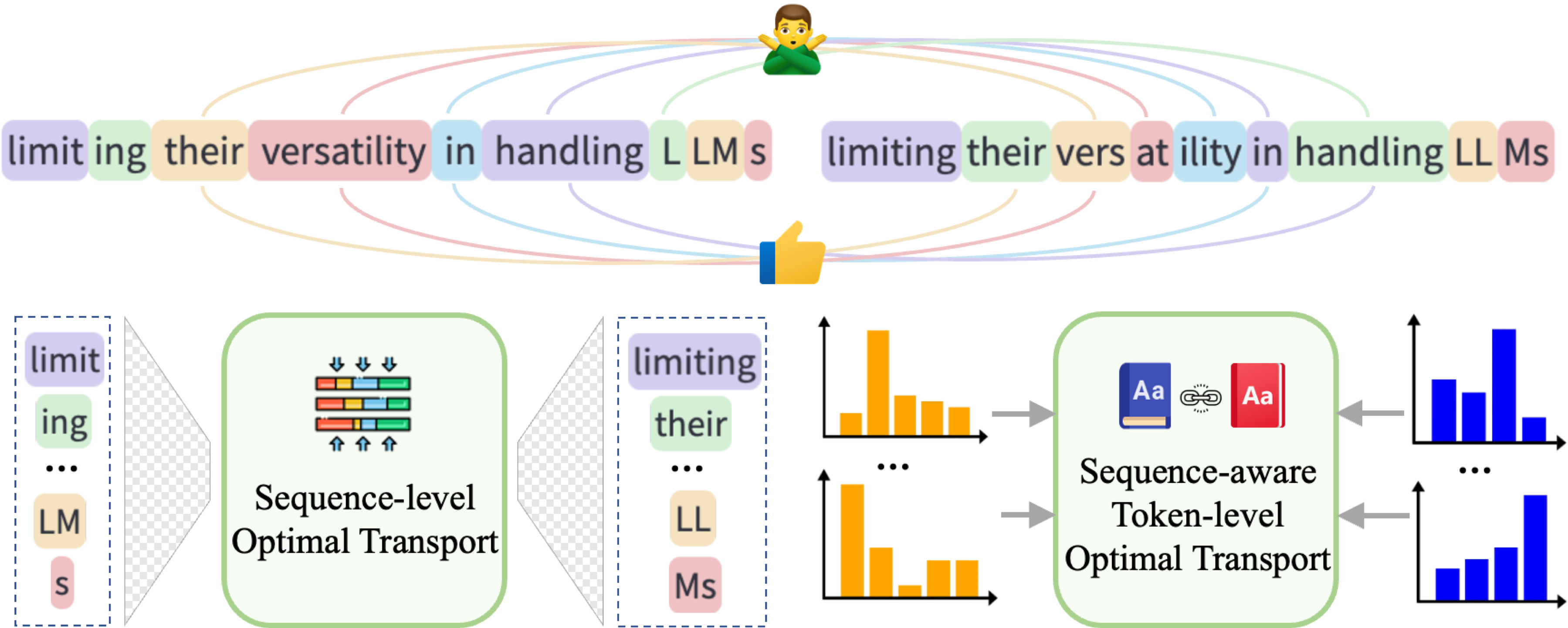}
\caption{
An illustration of vocabulary mismatch resulting from cross-tokenizer discrepancies. Unlike strict token-wise distillation methods that may lead to token misalignment, we employ sequence-level and sequence-aware token-level optimal transport to facilitate effective knowledge transfer.
}
\label{fig1}
\end{center}
\end{figure}

Cross-tokenizer knowledge distillation (CTKD) refers to the process of transferring knowledge between models that use different tokenizers (see Figure~\ref{fig1}).
It is crucial to ensure compatibility for applications such as multi-teacher knowledge transfer, where the student model learns from multiple teacher models with potentially different tokenization schemes.
However, most existing KD methods rely on divergence measures such as Kullback–Leibler (KL) divergence~\cite{kd,ckd,kd20241,kd20243}, reverse KL (RKL) divergence~\cite{rkl,minillm}, and Jensen–Shannon (JS) divergence~\cite{f-div,js1,js2}.
These measures require a strict point-by-point correspondence across dimensions between the student and teacher, necessitating the use of the same tokenizer and consistent vocabularies, which limits their applicability when different tokenizers are involved.

Very few studies notice such deficiency in directly applying existing KD techniques on LLMs,
for the simple reason that most KD methods are developed for few mainstream open-source models. 
ULD~\cite{uld}, the first attempt ever to tackle this issue, aligns the distributions of individual tokens between the teacher and the student using token-wise optimal transport (OT).
However, ULD focuses solely on the internal information of individual tokens
without considering the global context for robust 
matching.
Additionally, its reliance on zero padding 
introduces noise and hinders the effective use of logarithmic cost matrices.
DSKD~\cite{dual}, another token-wise alignment method, tries to transform the hidden states of one model to the space of another one bidirectionally via learnable projectors.
Despite its efforts in alignment for a unified output space,
DSKD fails to effectively leverage the distribution information as the transformed distribution often exhibits low accuracy.
Also, although these methods avoid strict dimensional correspondence, they assume a rigid token-by-token correspondence, which is often not the case in practice.



\begin{figure*}[tbp]
\begin{center}
\includegraphics[width=0.9\linewidth]{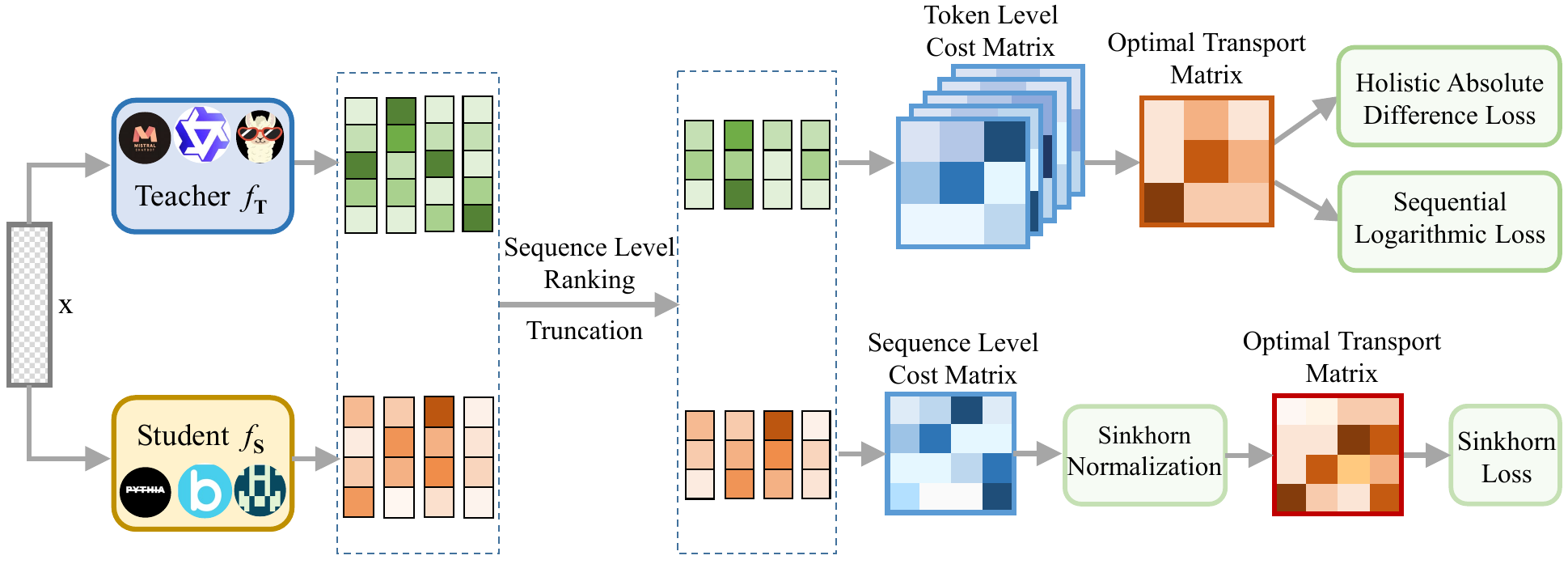}
\caption{
Illustration of our pipeline. MultiLevelOT computes sequence-aware token-level and sequence-level optimal transport distances between the output logits of the teacher and student models. This approach effectively transfers local and global information within the logits distribution, accommodating vocabulary differences and enabling cross-tokenizer distillation.
}
\label{framework}
\end{center}
\end{figure*}

To address these shortcomings
, we propose the Multi-Level Optimal Transport (MultiLevelOT) for cross-tokenizer knowledge distillation on LLMs.
Our method comprehensively measures the discrepancy between teacher and student logit distributions by calculating the optimal transport distance both within and across tokens in each sequence.
Such a dual-level approach ensures that both token-level and sequence-level relationships
are incorporated into the distillation process, effectively eliminating the need for dimensional or token-by-token correspondence.

At the token level, we jointly optimize all tokens within a sequence by
minimizing token-level discrepancies within the context of the entire sequence.
This is achieved by applying a sequence-level ranking process, which enables the same optimal transport plan for all tokens and effectively selects the important dimensions.
To eliminate noise from redundant dimensions,
we truncate the logits, focusing only on the most impactful logit dimensions for each sequence.
This truncation ensures that the teacher and student logits share a common support size, making each dimension meaningful and applicable for a logarithmic form cost matrix.
To capture both the fine-grained, token-wise nuances and the holistic, sequence-scale context view,
we employ two types of cost matrices:
one in the form of 
absolute difference and the other in the form of logarithm-based likelihood difference.
The absolute difference cost matrix captures the direct discrepancies in logits, providing a straightforward and interpretable measure of distance.
Conversely, the logarithmic cost matrix accounts for the relative differences
to offer a more nuanced and scalable measure.
It is particularly effective in handling logits with a wide range of magnitudes.



At the sequence level,
which has not been considered in previous studies, we utilize token-to-token OT distances to construct the sequence-level cost matrix. Since optimal transport automatically finds the corresponding relationships between tokens, this is particularly crucial for addressing token order misalignment caused by varying tokenization of long words across different tokenizers.
Unlike token-level transport, which deals with individual logit values,
sequence-level transport requires calculating the optimal transport between vectors of tokens.
Given the computational intensity of directly computing the Wasserstein distance for this purpose, we employ the Sinkhorn distance as an efficient approximation. 
This approach retains the benefits of the Wasserstein distance while significantly reducing computational complexity.
Importantly, we achieve all the improvements without introducing additional modules or modifying output formats specific to NLP tasks.

Extensive experiments are conducted in view of 1) \textbf{comparability},
2) \textbf{validity},
and 3) \textbf{generalizability}.
For comparability,
we test our method on different tasks
under both labeled and unlabeled distillation settings.
Our method consistently outperforms the state-of-the-art CTKD methods.
For validity,
we provide a comprehensive analysis through ablation studies and hyper-parameter tuning,
which corroborate the effectiveness of each component.
For generalizability,
the proposed method is validated on different students across families,
architectures,
and sizes.
We also experiment with diverse teachers
to demonstrate its robustness across various model choices. In summary, our contributions are:
\begin{itemize}
\item We propose the MultiLevelOT, a cross-tokenizer knowledge distillation approach that leverages both sequence-aware token-level and sequence-level optimal transport for comprehensive distribution matching.
\item 
We enhance the robustness of our method by jointly optimizing all tokens and using varied cost matrices, effectively capturing both global and local information.
\item We demonstrate the superiority of MultiLevelOT over existing methods through extensive experiments, validating its comparability, validity, and generalizability.
\end{itemize}

\section{Related Work}
\subsection{Knowledge Distillation}
Knowledge distillation (KD) is 
proposed to transfer the intrinsic knowledge from a teacher model to a student model by approximating the teacher's soft targets, such as output logits and intermediate representations. Cross-Tokenizer KD extends this traditional framework to scenarios involving different tokenizers, each with distinct vocabularies, which is crucial for LLM distillation. 
Various KD methods have been explored, ranging from logit-based distillation to representation-based distillation. These methods typically employ divergence measures like KL divergence 
\cite{kd,kd20241,kd20243,metadistill,reaugkd,mgskd},
RKL \cite{rkl,minillm,kd20242}, and JS divergence \cite{f-div,js1,js2}. These 
measures compute discrepancies on each dimension, requiring a one-to-one correspondence between teacher and student logit dimensions.
SinKD~\cite{sinkd,cui2024sinkd} addresses the limitations of these traditional measures by using the Sinkhorn distance. 
However, its approach still requires dimensional correspondence in the cost matrix.
In cross-tokenizer distillation, such dimensional correspondence is 
absent, making these methods inapplicable.

To overcome this challenge,
both ULD~\cite{uld} and DSKD~\cite{dual} propose promising solutions for token-wise alignment.
ULD measures token-wise OT distance between the logits of the student and teacher models, eliminating the dependency on dimensional correspondence.
DSKD attempts to transform the hidden states of one model to that of another by training projectors, but the transformed distribution often exhibits low accuracy.
Comparatively,
the proposed method differs in the following aspects:
1) ULD only considers local information while neglecting global distributional properties.
Its padding approach,
more like an ad-hoc 
brutal tactic,
limits it to
a singular cost matrix.
In contrast,
we stem from the token-level and sequence-level perspectives and deduce different forms of cost matrices
for lexical and semantic alignment.
2) While DSKD relies on traditional divergence measures, which suffer from issues like mode-averaging and mode-collapsing~\cite{sinkd}, we employ the Sinkhorn distance 
to fully capture the geometric characteristics of logit distributions.
In addition,
we do not explicitly enforce cross-model space mapping because such dual-space projection lacks semantic interpretability and thereafter hinders sequence comprehension.
3) Both ULD and DSKD assume a rigid token-by-token correspondence, which is often impractical. Our approach uses sequence-level OT, which automatically identifies corresponding relationships between tokens, thereby eliminating the need for strict token correspondence.


\subsection{Optimal Transport}
Optimal transport (OT) theory offers a robust mathematical framework for comparing probability distributions by calculating the minimal cost required to transform one distribution into another. The Wasserstein distance, a pivotal concept in OT, quantifies this cost and excels in capturing the geometric structure of distributions~\cite{wd,tnnls3}. This metric has been instrumental across various domains, including causal discovery~\cite{causal,causal1}, image generation~\cite{wgan1,wgan2,wgan3}, unsupervised learning~\cite{un1,un2,un3}, and reinforcement learning~\cite{re1,re3,re4}.

While the Wasserstein distance may be simplified in some low-dimensional cases, it can be computationally intensive in other scenarios. To address this, the Sinkhorn distance has been proposed as an approximation, which introduces an entropy regularization term to the OT problem, making it more tractable~\cite{sinkhornbase}. 
This approach has demonstrated success in diverse applications such as machine translation~\cite{li2023improving}, domain adaptation~\cite{nguyen2022improving,tnnls4}, classification~\cite{liu2023bilaterally},  and teacher model selection~\cite{more1,more2}.

Our approach employs both token-level and sequence-level OT for cross-tokenizer knowledge distillation. This dual-level OT captures global and local information, enhancing geometry information transfer and model efficacy.

\section{Methods}
\subsection{Problem Statement}
Given a sample $\mathbf{x}$ and its ground-truth label $\mathbf{y}$, 
the output logits with softmax activation $\sigma_{\tau}$ from the teacher $f_\textbf{T}$ and the student $f_\textbf{S}$ are  $\mathbf{t}\in\mathbb{R}^{T\times m}$ and $\mathbf{s}\in\mathbb{R}^{T\times n}$, respectively:
\begin{equation}
    \mathbf{t}=\sigma_{\tau}(f_\textbf{T} (\mathbf{x})),\quad \mathbf{s}=\sigma_{\tau}(f_\textbf{S} (\mathbf{x})),
    \label{eq:3}
\end{equation}
where $\tau$ represents the temperature parameter, $m$ and $n$ denote the dimensions of the teacher and student output vocabularies, respectively, and $T$ is the total number of tokens in the generated sequence. 
We denote the $i$-th dimension of the teacher and student logits for the $t$-th token as $\mathbf{t}_i(t)$ and $\mathbf{s}_i(t)$, respectively.
Our objective is to minimize the optimal transport distance between the distributions of the teacher's and student's outputs for knowledge transfer. In scenarios where the ground-truth label is unavailable, 
we use teacher-generated text as a substitute.

\subsection{Reconstructing optimal transport in ULD}
ULD~\cite{uld} leverages OT to address the challenge of cross-tokenizer knowledge distillation. To ensure equal support size between the teacher and student distribution spaces, 
ULD pads the smaller vocabulary with zero values, matching the larger size max($m,n$).
The ULD loss is then computed by summing the token-wise Wasserstein distances. 
The OT distance for the 
$t$-th token is defined as:
\begin{equation}
    \min_{\mathbf{P}(t)} \sum_{i=1}^{\max(m,n)}  \sum_{j=1}^{\max(m,n)}\mathbf{P}_{ij}(t)\mathbf{C}_{ij}(t), 
\end{equation}
where $\mathbf{P}$ is the optimal transport matrix and $\mathbf{C}$ is the cost matrix. ULD asserts that
each transport cost is equal to 1 and applies the following constraints on $\mathbf{P}$:
\begin{equation}
    \sum_i\mathbf{P}_{ij}(t)=\mathbf{s}_j(t) \quad \forall j,t, \sum_j\mathbf{P}_{ij}(t)=\mathbf{t}_i(t) \quad \forall i,t.
\end{equation}
However, the original formulation lacks flexibility.
We propose a more adaptable reformulation by setting
$\mathbf{C}_{ij}=\left|\mathbf{t}_i(t) -\mathbf{s}_j(t)\right|$ and using these constraints:
\begin{equation}
    \sum_i\mathbf{P}_{ij}(t)=1 \quad \forall j,t, \quad \quad \sum_j\mathbf{P}_{ij}(t)=1 \quad \forall i,t.
\end{equation}
Both formulations yield the same optimal transport distance:
\begin{equation}
 \mathcal{L}_{\text{ULD}} = \sum_{t=1}^T\sum_{i=1}^{\max(m,n)} \left|\mathbf{t}_{\text{TR},i}(t)-\mathbf{s}_{\text{TR},i}(t)\right|,
\end{equation}
where $\mathbf{s}_{\text{TR}}(t)$ and $\mathbf{t}_{\text{TR}}(t)$ are the token-wise ranked logits of the student and teacher, respectively:
\begin{equation}
\begin{aligned}
\mathbf{s}_{\text{TR}}(t) &= \mathbf{s}\left[ \operatorname{argsort} \left(  \mathbf{s}(t), \text{descending} \right) \right] \\
\mathbf{t}_{\text{TR}}(t) &= \mathbf{t}\left[ \operatorname{argsort} \left( \mathbf{t}(t), \text{descending} \right) \right].
\end{aligned}
\end{equation}
By reconstructing equivalent optimal transport problems, we can design various cost matrices and extend token-wise optimal transport distance to multi-level optimal transport.


\subsection{Multi-Level Optimal Transport}
Instead of considering each token independently, our method jointly optimizes all tokens within a sequence through sequence-aware multi-level OT, effectively aligning the distributions of teacher and student output logits. 
The primary objective is to minimize the sum of token-level and sequence-level costs using an optimal transport plan $\mathbf{P}$:
\begin{equation}
\min_\mathbf{P}  \sum_{i=1}^m \sum_{j=1}^{n} \mathbf{P}_{ij} \sum_{t=1}^T\mathbf{C}_{ij}^{tok}(t) + \min_\mathbf{P} \sum_{i=1}^T \sum_{j=1}^T \mathbf{P}_{ij} \mathbf{C}_{ij}^{seq},     
\end{equation}
where $\mathbf{C}^{tok}$ and $\mathbf{C}^{seq}$ represent the token-level and sequence-level cost matrices, respectively. Specific mathematical formulations will be detailed in subsequent paragraphs.
The optimization is subject to the constraints:
\begin{equation}
    \sum_i\mathbf{P}_{ij}=1 \quad \forall j, \quad \quad \sum_j\mathbf{P}_{ij}=1 \quad \forall i.
\end{equation}
We model the token-level cost using both absolute difference and logarithmic forms, while the sequence-level cost is captured through the optimal transport distance between tokens. 
For token-level alignment, our optimization strategy integrates both global and local information by considering the entire sequence within the optimal transport process. 
The full pipeline is illustrated in Figure~\ref{framework}. 



\paragraph{Holistic Absolute Difference Loss}
\label{m1}
We define the first token-level cost matrix $\mathbf{C}_{ij}^{tok}(t)$ using the absolute difference between logits:
$\mathbf{C}_{ij}^{tok}(t)=\left|\mathbf{t}_i(t) -\mathbf{s}_j(t)\right|$, so that the Wasserstein distance can be obtained by solving this optimization problem:
\begin{equation}
\min_\mathbf{P} \sum_{t=1}^T \sum_{i=1}^m \sum_{j=1}^{n} \mathbf{P}_{ij} \left|\mathbf{t}_i(t) -\mathbf{s}_j(t)\right|.
\end{equation}
While ULD employs a separate optimal transport matrix for each token, leading to inconsistent dimensional relationship, our approach ensures robustness by performing sequence-level ranking across all logits within a sequence. This allows us to use a single optimal transport matrix for all tokens, ensuring consistent dimensional ordering within each token $t$.
Our sequence-level ranking process is defined as follow:
\begin{equation}
\mathbf{t}_{\text{SR}} = \mathbf{t}\left[ \operatorname{argsort} \left( \sum_{t=1}^{T} \mathbf{t}(t), \text{descending} \right) \right], \quad
\mathbf{s}_{\text{SR}} = \mathbf{Q}\mathbf{s}, 
\end{equation}
where 
$\mathbf{Q}=\mathbf{Q}^{*}$ is a permutation matrix 
used to match the dimensions of $\mathbf{s}$ with the corresponding dimensions of $\mathbf{t}_{\text{SR}}$ at the sequence level, satisfying:
\begin{equation}
\mathbf{Q}^{*} = \underset{\mathbf{Q}}{\text{argmin}} \sum_{t=1}^T \sum_{i=1}^m  \left| \mathbf{t}_{\text{SR},i}(t) - [\mathbf{Qs}(t)]_i \right|.
\end{equation}
To ensure the consistency of the support size, allow for a logarithmic cost matrix,
prevent mode-averaging, and reduce noise from unlikely words,
we conduct the top-k truncation as follows:
\begin{equation}
 \mathbf{s}_{\text{SR,Tr}}(t) = \mathbf{s}_{\text{SR}}(t)[:k], 
 \quad \mathbf{t}_{\text{SR,Tr}}(t) = \mathbf{t}_{\text{SR}}(t)[:k], 
\end{equation}
where $[: k]$ denotes the slicing operation for choosing the top-k elements of the vector.
Then the optimization problem can be reformulated as:
\begin{equation}\label{eq:optimaltruncate}
\min_\mathbf{P} \sum_{t=1}^T \sum_{i=1}^k \sum_{j=1}^{k} \mathbf{P}_{ij} \left|\mathbf{t}_{\text{SR,Tr},i}(t) -\mathbf{s}_{\text{SR,Tr},j}(t)\right|.
\end{equation}
The optimal transport matrix to the above Eq.~\eqref{eq:optimaltruncate} is $\mathbf{P}^{*}=\mathbf{P}^{\text{HAD}}$,
where
$\mathbf{P}^{\text{HAD}}_{ij}$ is 1 only when $i=j$, and 0 otherwise. The absolute difference loss, 
representing the solution to this optimization problem, is then computed as:
\begin{equation}
 \mathcal{L}_{\text{HAD}} = \sum_{t=1}^T\sum_{i=1}^k \left|\mathbf{t}_{\text{SR,Tr},i}(t)-\mathbf{s}_{\text{SR,Tr},i}(t)\right|.  
\end{equation}
In the following text, all instances of $\mathbf{t}^k $ and $\mathbf{s}^k$ refer to $\mathbf{t}_{\text{SR,Tr}}$ and $\mathbf{s}_{\text{SR,Tr}}$, respectively.




\paragraph{Sequential Logarithmic Loss}
\label{m2}
For the token-level cost matrix, in addition to the absolute difference, we also incorporate a logarithmic form:
$\mathbf{C}_{ij}^{tok}(t)=-\mathbf{t}_i(t)\log\mathbf{s}_j(t)$. 
We apply the previously mentioned top-k truncation, which ensures that no zero-value elements are present in the student logits, thus making this cost matrix meaningful and effective.
Given that each dimension is equally important, the optimization problem for computing the Wasserstein distance can be formulated in a sequence-level ranked order:
\begin{equation}
\min_\mathbf{P} \sum_{t=1}^T \sum_{i=1}^k \sum_{j=1}^{k} -\mathbf{P}_{ij} \mathbf{t}^k_i(t)\log\mathbf{s}^k_i(t).
\end{equation}
The optimization objective is minimized by the sequential transfer between logit dimensions, making the optimal transport matrix $\mathbf{P}^{\text{SL}}$ equivalent to $\mathbf{P}^{\text{HAD}}$. Consequently, the loss function is defined as:
\begin{equation}
    \mathcal{L}_{\text{SL}} = -\sum_{t=1}^T\sum_{i=1}^k \mathbf{t}^k_i(t)\log\mathbf{s}^k_i(t).
\end{equation}

\paragraph{Sinkhorn Distance Loss}
We employ the optimal transport distance between tokens to measure pairwise differences between the $i$-th and $j$-th tokens in a sequence, constructing the sequence-level cost matrix  
$\mathbf{C}\in\mathbb{R}^{T\times T}$ with entries $\mathbf{C}_{ij}^{seq}=\sum_{l=1}^k \sum_{q=1}^{k} \mathbf{P}^{\text{HAD}}_{lq} \left|\mathbf{t}^k_l(i) -\mathbf{s}^k_q(j)\right|$.
Following SinKD~\cite{sinkd,cui2024sinkd}, we
use Sinkhorn distance as an efficient approximation for Wasserstein distance, 
retaining its benefits while significantly reducing computational costs for online distillation. 
The Sinkhorn distance is based on the relaxed formulation of an OT plan with entropy regularization. The OT plan $\mathbf{P}^{\lambda}$ is obtained by minimizing:
\begin{equation}
\mathbf{P}^{\lambda}=\underset{\mathbf{P}}{\text{argmin}}\sum_{i=1}^T \sum_{j=1}^T \mathbf{P}_{ij} \mathbf{C}_{ij}
- \lambda h\left(\mathbf{P}\right),
\end{equation}
where $h(\mathbf{P})$ is the entropy of the matrix $\mathbf{P}$, $\lambda>0$ is the entropy regularization weight.
To solve this iteratively, we construct the kernel matrix $\mathbf{K}^0\in\mathbb{R}^{T\times T}$ by applying the Gaussian kernel to $\mathbf{C}$
with the parameter $\lambda$:
\begin{equation}
\label{eq:initK}
    \mathbf{K}^0=\exp\left(-\frac{\mathbf{C}}{\lambda}\right).
\end{equation}
The OT plan $\mathbf{P}^{\lambda}$
is then derived through sequence-level Sinkhorn normalization, using iterative updates on $\mathbf{K}$:
\begin{equation}
\label{eq:updateK2}
\mathbf{\widehat{K}}^{i}\leftarrow \mathbf{K}^{i-1} \oslash \left( \mathbf{K}^{i-1}\mathbf{1}_{b}\mathbf{1}_{b}^\top \right), 
\mathbf{K}^i\leftarrow \mathbf{\widehat{K}}^{i} \oslash \left( \mathbf{1}_{b}\mathbf{1}_{b}^\top\mathbf{\widehat{K}}^{i} \right).
\end{equation}
For simplicity,
irrelevant constants are excluded from the equations.
After a pre-determined number of iterations $N$, the OT matrix is 
obtained as $\mathbf{P}^{\lambda}=\mathbf{K}^{N}$.
The sequence-level optimal transport distance loss is then computed as: 
\begin{equation}\label{eq:sinkdloss}
\mathcal{L}_{\text{SD}}=\left<\mathbf{P}^{\lambda},\mathbf{C}\right>=\sum_{i=1}^T \sum_{j=1}^T{\mathbf{K}^{N}_{i,j}\mathbf{C}_{i,j}}.
\end{equation}

\paragraph{Total Loss}
We combine the Cross-Entropy (CE) loss with the weighted holistic absolute difference loss, sequential logarithmic loss, and Sinkhorn distance loss for distillation. For a sequence of $T$ tokens, the total loss
is defined as:
\begin{equation}\label{eq:totalloss}
\mathcal{L}=\sum_{t=1}^T\mathcal{L}_{\text{CE}}\left(\mathbf{y}(t),\mathbf{s}(t)\right)+\alpha(\mathcal{L}_{\text{HAD}}+\beta\mathcal{L}_{\text{SL}}+\gamma\mathcal{L}_{\text{SD}}),
\end{equation}
where $\alpha$, $\beta$ and $\gamma$ are weights for each loss component. 






\section{Experiments}
\subsection{Experimental Settings}
\paragraph{Datasets.}
We evaluate our method on three representative tasks: an extractive QA task (QED)~\cite{qed}, a generative QA task (FairytaleQA)~\cite{fairtale}, and a summarization task (DIALOGSum)~\cite{dialogsum}. 
For evaluation,
we use the F1 score for the QED and the Rouge-LSum~\cite{rouge} for others.
More details are given in the appendix.

\begin{table}[tbp]
\footnotesize
\centering
\setlength{\tabcolsep}{0.5mm}
\begin{tabular}{lcccc}
\toprule
\textbf{Model} & \textbf{Method} & \textbf{\begin{tabular}[c]{@{}c@{}}QED\\ (F1)\end{tabular}} & \textbf{\begin{tabular}[c]{@{}c@{}}FairytaleQA\\ (Rouge-LSum)\end{tabular}} & \textbf{\begin{tabular}[c]{@{}c@{}}DIALOGSum\\ (Rouge-LSum)\end{tabular}} \\ \midrule
LLaMA2-7B & Few-Shot & 61.68 & 50.90 &37.75  \\ \midrule
\multirow{6}{*}{OPT-350M} & Origin & 12.46 &11.16 &14.02 \\ 
& SFT & 55.71 &46.04  &35.59  \\ 
& SeqKD & 49.61 &39.19  &30.71  \\ 
& MinED &56.03&46.11& 35.82\\
& ULD & 56.76 &45.82  &36.05  \\ 
& Ours &\bf 58.97 &\bf 46.96 &\bf 37.61\\ \midrule
\multirow{6}{*}{Pythia-410M} & Origin & 22.87 &15.14 &4.41 \\ 
& SFT & 59.03 &47.23  & 36.06 \\ 
& SeqKD & 51.12 &39.78  &31.57  \\ 
& MinED &59.21&47.31 &35.97 \\
& ULD & 59.71 & 47.81 & 36.07 \\ 
& Ours &\bf 61.79 &\bf 49.10 &\bf 37.45\\ \midrule
\multirow{6}{*}{Bloomz-560M} & Origin & 47.67 &43.47 &11.82 \\ 
& SFT & 60.48 &49.07  &36.52  \\ 
& SeqKD &52.33  &45.68  &31.83  \\ 
& MinED &60.52&49.10& 36.39\\
& ULD & 61.22 &49.87  &36.40  \\ 
& Ours &\bf 62.58 &\bf 50.94 &\bf 37.68\\ \midrule
\multirow{6}{*}{Average} & Origin &27.67 &23.25 &10.08 \\ 
& SFT & 58.41 &47.45  &36.05  \\ 
& SeqKD &50.99   &41.55  &31.37  \\
& MinED &58.58&47.47 &36.06 \\
& ULD & 59.30 &47.83  & 36.17 \\ 
& Ours &\bf 60.99 &\bf 49.00 &\bf 37.58\\ \bottomrule
\end{tabular}
\caption{Performance of the students in labeled distillation.
Both the teacher and ground-truth provide supervision.
}
\label{main1}
\end{table}

\begin{table}[tbp]
\footnotesize
\centering
\setlength{\tabcolsep}{0.5mm}
\begin{tabular}{lcccc}
\toprule
\textbf{Model} & \textbf{Method} & \textbf{\begin{tabular}[c]{@{}c@{}}QED\\ (F1)\end{tabular}} & \textbf{\begin{tabular}[c]{@{}c@{}}FairytaleQA\\ (Rouge-LSum)\end{tabular}} & \textbf{\begin{tabular}[c]{@{}c@{}}DIALOGSum\\ (Rouge-LSum)\end{tabular}} \\ \midrule
LLaMA2-7B & Few-Shot & 61.68 & 50.90 & 37.75 \\ \midrule
\multirow{4}{*}{OPT-350M} & Origin & 12.46 &11.16 &14.02 \\ 
& Raw Text & 49.61 &39.19  &30.71 \\ 
& ULD &50.71  &39.86  &32.03  \\ 
& Ours & \bf 51.96 &\bf 40.68 &\bf 36.88 \\ \midrule
\multirow{4}{*}{Pythia-410M} & Origin & 22.87 &15.14 &4.41 \\ 
& Raw Text & 51.12 & 39.78  &31.57  \\ 
& ULD &52.09  &40.69  &34.15  \\ 
& Ours &\bf 53.56 &\bf 41.28 &\bf 36.52 \\ \midrule
\multirow{4}{*}{Bloomz-560M} & Origin & 47.67 &43.47 &11.82 \\ 
& Raw Text & 52.33 &45.68  &31.83  \\ 
& ULD &53.02  &46.72  & 34.21 \\ 
& Ours &\bf 54.15 &\bf 47.88& \bf 37.10 \\ \midrule
\multirow{4}{*}{Average} & Origin & 27.67&23.25 &10.08 \\ 
& Raw Text &50.99 &41.55&31.37 \\ 
& ULD &51.94 &42.42 & 33.46\\ 
& Ours &\bf 53.22 &\bf 43.28 &\bf 36.83 \\ \bottomrule
\end{tabular}
\caption{
Performance of the students in unlabeled distillation.
The ground-truth is unavailable for supervision.
}
\label{main2}
\end{table}

\begin{table}[bp]
\footnotesize
\centering
\setlength{\tabcolsep}{1.5mm}
\begin{tabular}{cccccccccc}
\toprule
CE & AD & TR & SR & Tr & SL &SD & OPT & Pythia & Bloomz \\
\midrule
\checkmark &&&&&&&55.71&59.03&60.48 \\
\checkmark &\checkmark &\checkmark&&&&&56.76&59.71&61.22 \\
\checkmark &\checkmark&&\checkmark&&&&58.02&60.18& 61.56\\
\checkmark &\checkmark&&\checkmark&\checkmark&&&58.01&60.22&61.58 \\
\checkmark &\checkmark&&\checkmark&\checkmark&\checkmark&&58.17&61.10&61.87 \\
\checkmark &\checkmark&&\checkmark&\checkmark&&\checkmark&58.15&61.20&61.90 \\
\checkmark &&&\checkmark&\checkmark&\checkmark&\checkmark& 58.23& 61.17& 61.80 \\
\checkmark &\checkmark&&\checkmark&\checkmark&\checkmark&\checkmark&\bf 58.97&\bf 61.79&\bf 62.58 \\
\bottomrule
\end{tabular}
\caption{Ablation Study on QED across three students.}
\label{ablation}
\end{table}

\paragraph{Implementation details.}
We use four advanced teacher models: LLaMA2 7B Chat~\cite{LLaMA2}, Mistral3 7B Instruct~\cite{mistral}, Qwen 7B Chat~\cite{qwen1} and LLaMA3 8B Instruct~\cite{meta2024}. 
These models are chosen for their proficiency in few-shot learning 
and their unique vocabulary coverage
~\cite{gpt3}. 
For student models, we use a range of LLMs from various families and architectures,
including OPT 350M~\cite{opt}, Pythia 160M, Pythia 410M, Pythia 1B~\cite{pythia}, Bloomz 560M~\cite{bloomz}, and mT0 300M~\cite{bloomz}, initializing them with their pretrained weights. 
Following ULD~\cite{uld}, we set the learning rate $lr=1e-6$, $\alpha=0.15$, $\beta=0.1$. Additionally, we empirically set $\gamma=0.1$, $\tau_{\text{SL}}=1$, $\tau_{\text{SD}}=2$,
$\lambda=0.1$,
$N=20$ and $k=50$.
Discussions on the effects of key factors $N$, and $k$ are presented later. 
Although further tuning may enhance performance, we maintain a consistent set of hyper-parameters across all tasks to underscore the robustness of our approach. 
\paragraph{Baselines.}
Our experiments involve two settings: 
labeled distillation and unlabeled distillation.
Labeled distillation, commonly used in most distillation studies, involves supervision with ground-truth labels. In contrast, unlabeled distillation relies solely on the generated texts from the teacher as pseudo-targets~\cite{uld}.
For labeled distillation, we compare our approach against the following baselines: Supervised Fine-Tuning (SFT), Sequence-level KD (SeqKD) \cite{seqkd}, MinED \cite{mined}, and ULD~\cite{uld}.
SeqKD can be interpreted as a form of supervised fine-tuning using the teacher’s outputs, deriving knowledge exclusively from the teacher model.
MinED, which can align the logits 
using dynamic programming 
, is also included in our comparison.
For unlabeled distillation, we follow the ULD to adopt the same baselines.
In both settings, we use the official code and default hyper-parameters for each baseline to ensure a fair comparison.
We exclude DSKD~\cite{dual} from our comparison as it 
introduces additional modules whose increased learnable parameters may cause unfair comparison.

\subsection{Results and Discussions}

\begin{table}[tbp]
\small
    \centering
    \begin{tabular}{lccc}
    \toprule
   & OPT & Pythia & Bloomz \\
   \midrule
    w/o SD loss    &58.17&61.10&61.87 \\
    w token-level  SD loss    &58.32 & 61.22 & 61.95\\
    w sequence-level  SD loss  & \bf 58.97&\bf 61.79&\bf 62.58 \\
    \bottomrule
    \end{tabular}
    \caption{Comparison of token-level and sequence-level Sinkhorn distance loss on QED across three students.}
    \label{seqandtoken}
\end{table}

\begin{figure}[tbp]
\begin{center}
\includegraphics[width=0.48\linewidth]{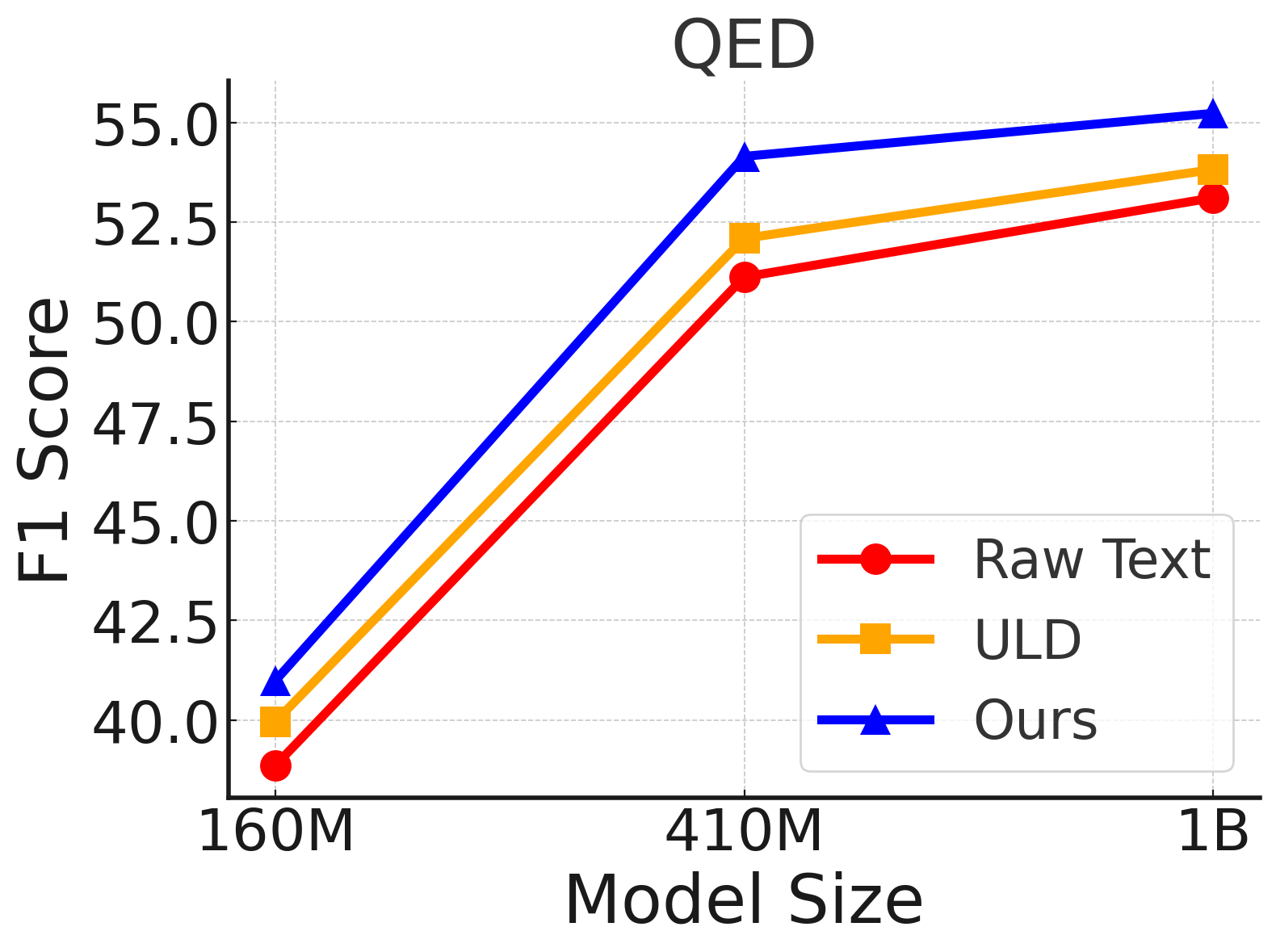}
\includegraphics[width=0.48\linewidth]{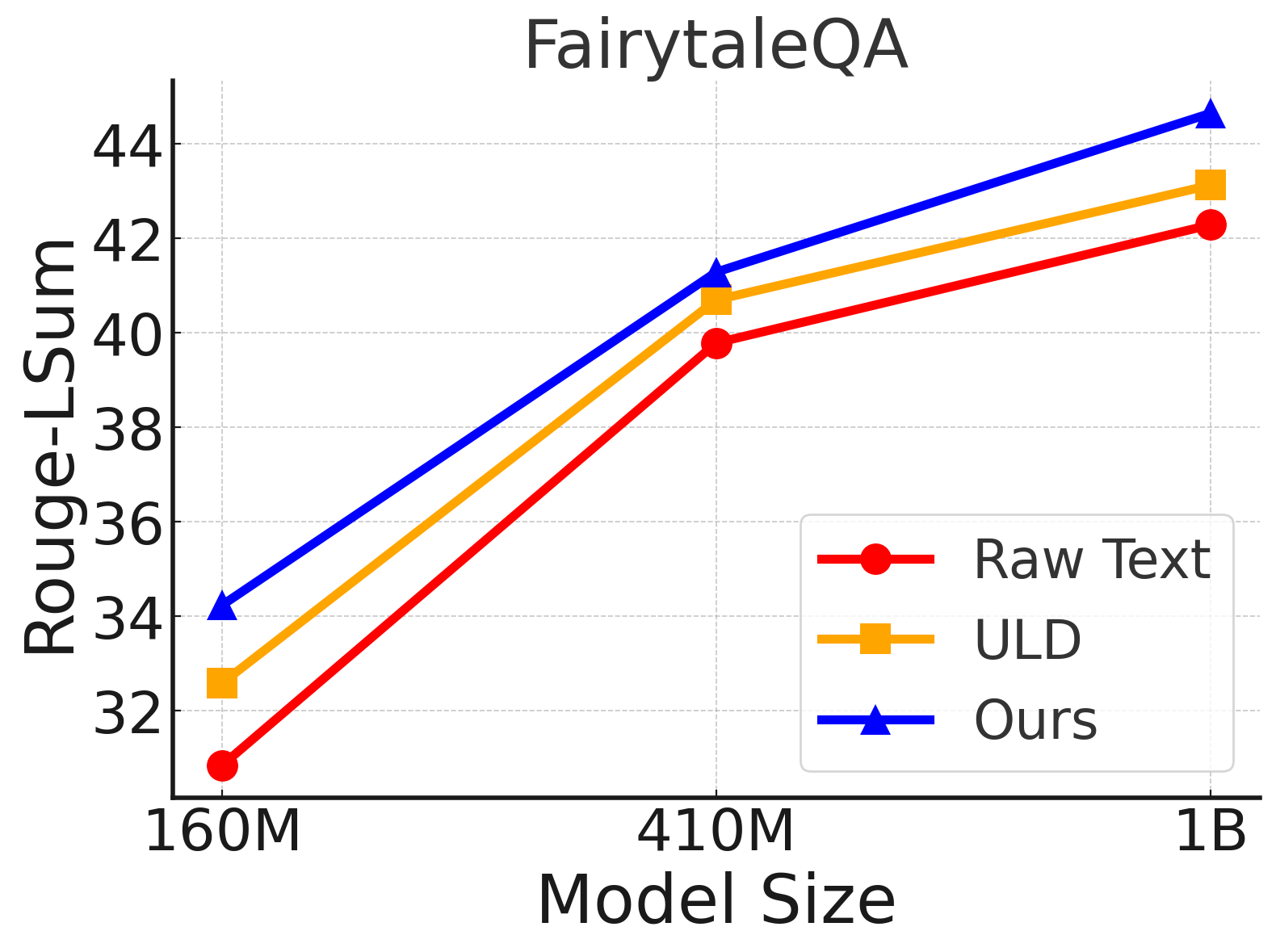}
\caption{
Performance at different student scales (Pythia 160M, 410M, and 1B) on QED and FairytaleQA.
}
\label{scale}
\end{center}
\end{figure}



\paragraph{Comparison with SOTA.}
Results on labeled distillation and unlabeled distillation are presented in Table~\ref{main1} and Table~\ref{main2}, respectively.
MultiLevelOT consistently outperforms all baseline methods across all datasets and student models. Notably, compared with ULD~\cite{uld}, MultiLevelOT reduces the performance gap between the student and the teacher by over 71\% in the QED task on labeled distillation. This improvement highlights the effectiveness of MultiLevelOT in bridging the performance gap by transferring sequence-level and sequence-aware token-level knowledge from the teacher to the student.
The superior performance of our approach is also attributed to the well-rounded design of the cost matrix. By employing diverse cost matrices, we facilitate effective geometry distribution information extraction and enhance the knowledge transfer process.

\begin{table}[tbp]
\small
\centering
\begin{tabular}{lccc}
\toprule
\textbf{Method} & \textbf{\begin{tabular}[c]{@{}c@{}}QED\\ (F1)\end{tabular}} & \textbf{\begin{tabular}[c]{@{}c@{}}FairtaleQA\\ (Rouge-LSUM)\end{tabular}} & \textbf{\begin{tabular}[c]{@{}c@{}}DIALOGSum\\ (Rouge-LSUM)\end{tabular}} \\ \midrule
Raw Labels  &34.96  &29.73  & 28.88 \\
ULD   &37.25  &31.52  &30.04  \\ 
Ours &\bf 41.37 &\bf 34.01 &\bf 33.01 \\ \bottomrule
\end{tabular}
\caption{Generalizability of MultiLevelOT in student architecture. Teacher: LLaMA, student: mT0-300M.}
\label{tab:mt0}
\end{table}

\begin{table}[tbp]
\small
\centering
 \begin{tabular}{lcccc}
    \toprule
   Method & LLaMA2& Mistral3& Qwen & LLaMA3\\
   \midrule
    Teacher     &61.68&64.03&62.16 &65.96\\
    Raw Text    &49.61 &51.24&51.21 &51.91\\
    ULD   &50.71 &52.08&52.89 &52.81\\
    Ours   &\bf 51.96  &\bf 52.96&\bf 53.99&\bf 54.38 \\
    \bottomrule
    \end{tabular}
\caption{Generalizability of MultiLevelOT across different teacher models on QED. Student : OPT-350M.}
\label{teacher}
\end{table}

\paragraph{Each components plays its role in MultiLevelOT.}
The ablation study on the QED task, as shown in Table~\ref{ablation}, demonstrates the critical role of each component in the MultiLevelOT framework. The baseline model, utilizing only cross-entropy (CE) loss, corresponds to the standard SFT. 
Adding the absolute difference (AD) and token-wise ranking (TR), as in ULD, provides a reference for improvement.
However, the key advancements come from our proposed components.
Integrating sequence-level ranking (SR) and truncation (Tr) with AD results in the 
\textbf{Holistic Absolute Difference Loss}, which shows significant gains by capturing both global and local geometrical information.
Incorporating the \textbf{Sequential Logarithmic Loss (SL)} further boosts performance, highlighting the value of various cost matrices in capturing different aspects of the distribution. Finally, integrating the \textbf{Sinkhorn Distance Loss (SD)} results in the best performance,
underlining the necessity of sequence-level knowledge for effective knowledge transfer.


\paragraph{Sequence-level Sinkhorn distance excels token-level Sinkhorn distance.}
Table~\ref{seqandtoken} demonstrates that sequence-level Sinkhorn distance outperforms token-level distance across all student models. The sequence-level approach captures the geometric properties of logit distributions more comprehensively, providing a robust framework for understanding global contextual relationships among tokens. In contrast, while token-level distance, akin to a Holistic Absolute Difference Loss with an added entropy term, enhances robustness and mitigates sparsity, it fails to fully encapsulate the overarching patterns of entire sentences.

\begin{table}[tbp]
\small
\centering
\begin{tabular}{lccccc}
\toprule
\textbf{Number of}   & \multirow{2}{*}{5} & \multirow{2}{*}{10} & \multirow{2}{*}{20} & \multirow{2}{*}{50} & \multirow{2}{*}{100}\\ 
    \textbf{Iterations $N$} &&&&& \\
\midrule
OPT-350M &58.26&58.52&58.97&59.02& 58.99\\
Pythia-410M &60.56&61.24&61.79&61.76& 61.78 \\
\bottomrule
\end{tabular}
\caption{Effect of $N$ on QED.}
\label{effectt}
\end{table}

\begin{table}[tbp]
\small
\centering
\begin{tabular}{lccccc}
\toprule
\textbf{Truncation}   & \multirow{2}{*}{5} & \multirow{2}{*}{20} & \multirow{2}{*}{50} & \multirow{2}{*}{100} & \multirow{2}{*}{1000}\\ 
    \textbf{Threshold $k$} &&&&& \\
\midrule
OPT-350M &58.54 &58.84 &58.97 &58.78 & 58.42\\
Pythia-410M &61.42 &61.50 &61.79 &61.40 &61.32 \\
\bottomrule
\end{tabular}
\caption{Effect of $k$ on QED.}
\label{effectk}
\end{table}

\paragraph{MultiLevelOT generalizes well on student LLMs across scales.}
We evaluate the impact of student LLMs' sizes on the efficacy of MultiLevelOT through a detailed analysis in an unlabeled distillation context. Using two diverse tasks, QED~\cite{qed} and FairytaleQA~\cite{fairtale}, as illustrated in Figure~\ref{scale}, we observe that MultiLevelOT consistently enhances the performance of student models across various scales. This improvement substantiates MultiLevelOT's advanced capability to effectively utilize optimal transport for knowledge distillation, clearly outperforming the ULD method~\cite{uld}.

\paragraph{Generalization of MultiLevelOT across student architectures.}
Since MultiLevelOT relies solely on logits in the distillation process, it can be applied to any architecture. In addition to decoder-only models, we also test it on the encoder-decoder model mT0~\cite{bloomz}. 
Results in Table~\ref{tab:mt0}
reveal significant performance enhancements, underscoring MultiLevelOT's flexibility and effectiveness across various architectural frameworks. 


\paragraph{Generalization of MultiLevelOT across teacher LLMs.}
An extensive evaluation of MultiLevelOT's performance with varying teacher LLMs is conducted, employing models including LLaMA2 7B Chat~\cite{LLaMA}, Mistral 7B Instruct~\cite{mistral}, Qwen 7B~\cite{qwen1}, and LLaMA3 8B Chat~\cite{meta2024}. As shown in Table~\ref{teacher}, MultiLevelOT consistently outshines its counterparts. This 
highlights MultiLevelOT’s robust capacity to leverage the distinct advantages of various teacher models.
\paragraph{$N$ as the number of Sinkhorn iterations.}
We analyze the impact of varying the number of Sinkhorn iterations ($N$) on model performance, as summarized in Table~\ref{effectt}. Increasing $N$ to 20 led to substantial improvements in F1 scores for both OPT-350M (58.97) and Pythia (61.79), underscoring the importance of adequate iterations for achieving convergence. Beyond this point, however, raising $N$ to 50 yields negligible performance gains, indicating a saturation threshold where additional iterations do not contribute further. This suggests that while sufficient iterations are necessary for convergence, excessive iterations offer diminishing returns and unnecessarily increase computational costs. 

\paragraph{$k$ as the number of truncation threshold.}
Table~\ref{effectk} illustrates the effect of the truncation threshold ($k$) on knowledge distillation for two student models, OPT-350M and Pythia-410M. Our findings demonstrate that $k=50$ is optimal for both models on the QED dataset. A smaller $k$ insufficiently captures the full sentence structure, weakening the Sinkhorn distance's ability to model high-dimensional geometric information, and thus limiting the student model's capacity to mimic the teacher’s logit distribution. Conversely, a larger $k$ introduces too many near-zero logit elements, adding noise and causing mode-averaging, which impairs the student's ability to distinguish critical information.

\section{Conclusion}
We propose MultiLevelOT for cross-tokenizer knowledge distillation that leverages both sequence-aware token-level and sequence-level optimal transport. Our method incorporates diverse cost matrices, using joint token optimization and Sinkhorn distance to provide a robust and comprehensive framework for KD. Extensive experiments demonstrate that MultiLevelOT consistently outperforms state-of-the-art cross-tokenizer KD methods across various NLP tasks. Moreover, our approach proves robust across different student model families, architectures, sizes, and teacher models, showcasing its versatility and broad applicability. 


\paragraph{Broader Impact} It is prospective to use our method for multi-teacher knowledge transfer, integrating knowledge from multiple teachers to enhance model performance.  Additionally, MultiLevelOT may be suitable for cross-language and multi-modal knowledge transfer, enabling robust alignment across different languages and data modalities.

\section{Acknowledgments}
This work was supported by National Natural Science Foundation of China under Contract 62021001, and the Youth Innovation Promotion Association CAS. It was also supported by GPU cluster built by MCC Lab of Information Science and Technology Institution, USTC, and the Supercomputing Center of the USTC.

\bibliography{aaai25}

\end{document}